# BRAINSTORMING: Consensus Learning in Practice

Dariusz Plewczynski
ICM, Interdisciplinary Centre for Mathematical and Computational Modelling,
University of Warsaw,
Pawinskiego 5a Street, 02-106 Warsaw, Poland

e-mail: D.Plewczynski@icm.edu.pl

**Keywords**: machine learning, consensus, meta-learning, bioinformatics, chemoinformatics, brainstorming, neural networks, support vector machines, decision trees, random forest, genetic algorithms, nearest neighbours, trend vectors

**Abstract**: We present here an introduction to Brainstorming approach, that was recently proposed as a consensus meta-learning technique, and used in several practical applications in bioinformatics and chemoinformatics. The consensus learning denotes heterogeneous theoretical classification method, where one trains an ensemble of machine learning algorithms using different types of input training data representations. In the second step all solutions are gathered and the consensus is build between them. Therefore no early solution, given even by a generally low performing algorithm, is not discarder until the late phase of prediction, when the final conclusion is drawn by comparing different machine learning models. This final phase, i.e. consensus learning, is trying to balance the generality of solution and the overall performance of trained model.

# 1 INTRODUCTION

A novel meta-approach emerging in bioinformatics is called *consensus learning*. Here, one trains an ensemble of machine learning algorithms using different types of input training data representations. Then all solutions are gathered, and the consensus is build between them. Therefore no early solution, given even by a generally low performing algorithm, is not discarder until the late phase of prediction, when the final conclusion is drawn by comparing different machine learning models. The final phase, i.e. consensus learning, is trying to balance the generality of solution and the overall performance of trained model.

Although quite a few identifiers have been developed in this regard through various approaches, such as clustering, support vector machine (SVM), artificial neural network (ANN), or K-nearest neighbor (KNN), and many others classifiers the way they operate the identification is basically individual. Yet, the proper approach usually take into account the opinions from several experts rather than rely on only one when they are making critical decisions. Likewise, a sophisticated identifier should be trained by several different modes. This is the core idea of consensus learning that will be described in this manuscript, which applications were described previously for selected types of machine learning methods such as clustering [1-4], support vector machine [5-11], artificial neural networks [9, 12, 13], or K-nearest neighbor [5, 7, 12, 14] including our own findings [15-20]. Bagging and boosting are previously known meta-learning techniques had a wide array of applications as recapitulated in various manuscripts [21-29].

The consensus learning approach is similar to other ensemble methods, yet differently from bagging (combines many unstable predictors to produce a ensemble stable predictor), or boosting (combines many weak but stable predictors to produce an ensemble strong predictor), it focuses of the use of heterogeneous set of algorithms in order to capture even remote, weak similarity of the predicted sample to the training cases. The actual treatment of the input training data is also different from the previous approaches. The ultimate goal of machine learning is to discover the relationships between the variables of a system (input, output and hidden) from direct samples of the system. Most methods assumes single representation of training data. Here one builds the set of multiple hypotheses by manipulate the training examples, input data points, target output (the class labels) of the training data and by introducing randomness into the training data representation.

Generally there are two competing philosophies in supervised learning, where goal is to minimize the probability of model errors on future data. A single model approach is trying to build a single good model: either not using Occam's razor principle (Minimax Probability Machine, trees, Neural Networks, Nearest Neighbor, Radial Basis Functions) or those based on Occam's razor models that select the best model as the simplest one (Support Vector Machines, Bayesian Methods, other kernel based methods such as Kernel Matching Pursuit). An ensemble of models states that a good single model is difficult to compute, so it tries to build many of those and combine them. Combining many uncorrelated models produces better predictors as was observed in models that don't use randomness or use directed randomness (Boosting, Specific cost function, Gradient Boosting, a boosting algorithm derivative for any cost function), or in models that incorporate randomness (Bagging, Bootstrap Sample: Uniform random sampling with replacement, Stochastic Gradient Boosting, Random Forests, or by inputs randomizations for splitting at tree nodes).

Based on these experience and traditional learning approaches the consensus learning seems to be the method of choice where complex rules emerges from input training data, and the adaptivity of learning rules is of crucial importance. An consensus of different classifiers often outperforms a single classifier: a learning algorithm searches the hypothesis space to find the best possible hypothesis. When the size of training set is small, a number of hypotheses may appear to be optimal. An ensemble will average the hypotheses reducing the risk of choosing the wrong one. In addition most classifiers perform a local search often getting stuck in local optima; multiple starting points provide a better approximation to the unknown function. Finally a single classifier may not be able to represent the true unknown function. A combination of hypotheses, however, may be able to represent this function.

## 2 CONSENSUS OF MACHINE LEARNING METHODS

We present here a novel approach in machine learning namely *brainstorming*. Based on ordinary dictionary the brainstorming is a method of solving problems in which all the members of a group suggest ideas and then discuss them (a brainstorming session). In our approach we train an ensemble of machine learning algorithms using different types of input training data representations. In the second step all solutions are gathered and consensus (a general agreement about a matter of opinion) is build between them. Therefore no early solution, given even by a generally low performing algorithm, is not discarder until the late phase of prediction, when the final conclusion is drawn by comparing different machine learning models. This final phase, i.e. consensus learning, is trying to balance the generality of solution and the overall performance of trained model.

Our approach is similar to other ensemble methods, yet differently from bagging (combines many unstable predictors to produce a ensemble stable predictor), or boosting (combines many weak but stable predictors to produce an ensemble strong predictor), it focuses of the use of heterogeneous set of algorithms in order to capture even remote, weak similarity of the predicted sample to the training cases. The actual treatment of the input training data is also different from the previous approaches. The ultimate goal of machine learning is to discover the relationships between the variables of a system (input, output and hidden) from direct samples of the system. Most methods assumes single representation of training data. Here we build the set of multiple hypotheses by manipulate the training examples, input data points, target output (the class labels) of the training data and by introducing randomness into the training data representation.

The goal of this manuscript is to provide a general, theoretical framework for the general integration of results individual machine learning algorithms. In order to perform analytical analysis, we assume infinite, statistical ensemble of different ML methods. The global preference toward true solution can be described in our approach as the global parameter affecting all learners. Each learner (intelligent agent) performs training on available input data toward classification pressure described by the set of positive and negative cases. When the query testing data is analyzed each agent predicts the query item classification by "yes"/"no" decision. The answers of all agents are then gathered and integrated into the single prediction via majority rule in the field theoretical formulation. This view of the consensus as between various machine learning algorithms is especially useful for artificial intelligence, or robotic applications, where adaptive behavior given by the integration of results from ensemble of ML methods.

The consensus building between various machine learning algorithms, or various prediction outcomes, is similar to the weakly coupled statistical systems known from physics. The phase transitions can be observed in the system, the global new phase emerging when the system reaches a critical point in terms of its order parameter. Changes between phases of the system are induced by some external factors that can be modeled as a bias added to the local fields.

The model of meta-learning is based on several assumptions:

1. *Binary Logic*

   We assume the binary logic of individual learners, i.e. we deal with *N* learning agents that are modeled by different machine learning algorithms or different versions of the same ML method. Each agent, for the single prediction, holds one of two opposite states ("NO" or "YES"). These states are binary $\sigma_i = \pm 1$. In most cases the machine learning algorithms, such as support vector machines, decision trees, trend vectors, artificial neural networks, random forest, predict two classes for incoming data, based on previous experience in the form of trained models. The prediction of an agent answers single question: is a query data contained in class A ("YES"), or it is different from items gathered in this class ("NO").

2. *Disorder and random strength parameter*

   Each learner is characterized by two random parameters: $p_i = precission_i$ and $s_i = recall_i$ that describe the quality of predictions for individual agent for a given training data. Those values should be in principle averaged over different training data representations, different datasets used in training in order to make them data-independent. In present manuscript we assume that precise agents has high recall value $p_i = s_i$. In general the individual differences between agents are described as random variables with a probability density $\hat{p} = (p_i, s_i)$, with mean values $p = \sum \frac{p_i}{N}$ and $s = \sum \frac{s_i}{N}$.

3. *Learning preferences*

   The learning preference for each learning agent can be defined as the total learning impact $I_i$ that $i$th agent is experiencing from all other learners. This impact is the difference between positive coupling of those agents that hold identical classification outcome, relative to negative influence of those who share opposite state:

$$I_i = I_p\left(\sum_j \frac{t(p_j)}{N}\left(1-\sigma_i\sigma_j\right)\right) - I_s\left(\sum_j \frac{t(s_j)}{N}\left(1+\sigma_i\sigma_j\right)\right), \qquad (1)$$

where $t(p_i, s_j)$ is the strength scaling function. We take the strength scaling as $t(x) = x$.

*4.* The probability of success, *fuzzy logic*

The weighted majority-minority difference in the system is given by the equation:

$$m = \sum_j {(s_j + p_j)\sigma_j}/{N(s+p)}. \qquad (2)$$

The normalized value of $m$ describes the probability for the correct prediction, i.e. we assume here the vote rule. Each learner votes for the final prediction outcome, and all votes are gathered and the relative probability of correct answer is calculated, as given by the ensemble of learners.

*5.* Brainstorming: the procedure of *Consensus Learning*

The binary classification, i.e. consensus learning outcome $r$ of a prediction is given by the sign of weighted majority-minority difference for the whole system of individual learning agents:

$$r = sign(m) = sign\left(\sum_j {(s_j + p_j)\sigma_j}/{N(s+p)}\right). \qquad (3)$$

6. Presence of *noise*

The randomness of state change (phenomenological modeling of various random elements in the learning system, and training data) is given by introducing noise into dynamics:

$$r = sign\left(\sum_j\left({(s_j + p_j)\sigma_j}/{N(s+p)} + \beta h_j\right)\right), \qquad (4)$$

where and $\beta = 1/_{kT}$, and T represents "*temperature*" of the system. Temperature allows for simulating the competition between the deterministic outcome of consensus learning and stochastic nature of noise [30]. The random numbers $h_j$ is the site-dependent white noise, or one can select a uniform white noise, where for all agents $h_j = h$. In the first case $h_j$ are random variables independent for different agents and time instants, whereas in the second case $h$ are independent for different time instants. We assume here, that the probability distribution of $h_j$ is site, i.e. it has uniform statistical properties. The uniform white noise simulates the global bias affecting all agents, whereas site-dependent white noise describes local effects, such as prediction quality of individual learner etc.

The computational implementation of the above protocol is presented on Figure 1. Input objects and their descriptors are first represented by several methods in order to annotate them in the most enriched and efficient way. Then the resulting data is processed by feature decomposition in order to evaluate the statistical significance of each feature or representation, find some similarities between both objects and their features. Then training data prepared in such way is used for training several different machine learning methods (SVM, ANN, RF, DT and many others). The heterogeneous predictors classify the training data differently, therefore a consensus is needed for fusing their results. The prepared in the classification phase the MLcons meta-learner can further predict the class membership or selected features for a new objects, in prediction phase. The output contain reliability score, decision rules with significance of features used in prediction.

## 3 APPLICATIONS OF BRAINSTORMING

Based on experience and traditional learning approaches the consensus learning seems to be the method of choice where complex rules emerges from input training data, and the adaptivity of learning rules is of crucial importance. An consensus of different classifiers often outperforms a single classifier: a learning algorithm searches the hypothesis space to find the best possible hypothesis. When the training data is small, a number of hypotheses may appear to be optimal. An ensemble will average the hypotheses reducing the risk of choosing the wrong one. In addition most classifiers perform a local search often getting stuck in local optima; multiple starting points provide a better approximation to the unknown function. Finally a single classifier may not be able to represent the true unknown function. A combination of hypotheses, however, may be able to represent this function.

The first application of our approach was done for prediction of protein-protein interactions. Up to now, most of computational algorithms use single machine learning methods for analysis and prediction of protein-protein interactions [31-35], or statistical analysis of interacting patches of protein surfaces [36-39]. Our experience clearly supports the idea that each machine learning algorithm is performing better for selected types of training data [20, 40]. Some of them present very high specificity, others focus more on sensitivity. Sometimes one can have very large number of positives in training; on the other hand it is also common for some specific types of experiments that only few confirmed by experiments instances are known. In most cases the proper selection of negatives is not trivial. In the case of protein-protein interactions one should use rich variety of input data for training, such as sequences, short sequence motifs, evolutionary information, genomic context, enzymatic classification, known or predicted local or global structure of interacting proteins, and many others. Using this data one can apply various types of machine learning methods trained on the same set of positives and negatives, for example neural networks, support vector machine, random forest, decision trees, rough sets and others. The crucial step of meta-prediction is building consensus between those various prediction methods. Since systematic errors of multiple methods is usually randomly distributed the consensus approach can be used to select a common, probably the most accurate predictions [41]. A consensus method because of its easy parallelization can improve the accuracy of any single machine learning method without extending the time of prediction (the time needed is equal to the slowest used machine learning algorithm). The combination of various approaches done by Sen and Kloczkowski [42] provide the solid justification for this statement. They combine four different methods, such as data mining using Support Vector Machines, threading through protein structures, prediction of conserved residues on the protein surface by analysis of phylogenetic trees, and the Conservatism of Conservatism method of Mirny and Shakhnovich [43-45]. A consensus method predicts protein-protein interface residues by combining sequence and structure-based methods. Therefore the consensus approaches are one of most effective tools to handle prediction of protein-protein interactions on the whole proteomes level, what is the ultimate goal of system biology.

## 3 CONCLUSIONS

Meta-learning approach trains an ensemble of machine learning algorithms on the whole or different subset of all available training examples. The consensus gather all solutions and is trying to balance between them in order to maximize the prediction performance. No early solution, even provided by a generally low performing module, is not discarder until the late phase of prediction, when the final conclusion is drawn by comparing different machine learning classifiers. This final phase is focusing on balance the generality of solution and the overall performance of trained model. Early results shows, that the Brainstorming approach reaches higher performance than any single method used in consensus. This confirms reported results of other meta-learning approaches based on different versions of single machine learning algorithm or those that use a set of different ML.

The bioinformatics is enormously rich application field for mathematical methods. The complexity of scientific problems, large amount of heterogeneous biological data provide an excellent testgroud for machine learning approaches in real life context. In return, bioinformatics while using different theoretical methods, can also give back a serious advances in theoretical computational intelligence. Most computational approaches are based on comparative molecular similarity analysis of proteins with known and unknown characteristics. We have provided elsewhere an overview of publications which have evaluated different ML methods, especially focusing on ensemble and meta-learning approaches. Amongst the methods which are considered in this monograph are support vector machines, decision trees, ensemble methods such as boosting, bagging and random forests, clustering methods, neuronal networks, naïve Bayesian, data fusion methods, consensus and meta-learning approaches and many others. Therefore we do not report those applications here, focusing rather on theoretical foundations of consensus learning algorithm. Tom Dietterich at Cognet-02 stated that "the goal of machine learning is to build computer systems that can adapt and learn from their experience". We are therefore trying to mimic different performance tasks by applying different learning techniques trained on different input training data representations. Yet, the final conclusion raised by Niels Bohr is still valid: "Predicting is very difficult especially about the future". Even with recent advances of machine learning approaches the actual prediction phase is not perfect, and takes a lot of resources and time to perform. The ultimate goal of Artificial Intelligence studies, i.e. constantly evolving meta-learner that in real time accumulate the acquired information in the form of processed knowledge is still long way from the present state of research. Both, theoretical algorithms, and hardware resources (computers, or specialized accelerators) have to be improved in order to perform instant, rapid learning using different algorithms, when new input is presented to the system. Only then the "intelligent" system will be able to answer most of our expectations focusing on computational intelligence.

**ACKNOWLEDGEMENTS**

This work was supported by EC BioSapiens (LHSG-CT-2003-503265) and OxyGreen (KBBE-2007-212281) 6FP projects as well as the Polish Ministry of Education and Science (N301 159735, PBZ-MNiI-2/1/2005 and others).

**FIGURE 1.** The consensus learning protocol. Input objects are characterized by the set of descriptors, in most cases by the vectors of real or binary numbers. In the case of proteins typically amino acids sequence string and/or its 3D structure (positions of all atoms in Cartesian space) are used for describing proteins. The set of input objects is then passed to a set of computational tools in order to enrich the information used in training by external sources. In the case of protein sequences its 3D structural models (predicted by structure prediction web servers), physical or chemical properties of each amino acid of the sequence, set of homological sequences (identified by PSI-BLAST or other methors), biological annotations that can be found in external databases and processed by text mining techniques, and many others. Then the resulting data is stored in SQL database. All training objects, their description and additional annotations are then processed by feature decomposition module in order to evaluate the statistical significance of each descriptor or representation, find some similarities between objects, their features, annotations. In that way algorithm prepare a set of independent training datasets, in order to probe different representations of training data. All datasets are then used for training several different machine learning methods (SVM, ANN, RF, DT and many others). Each classifier learn separately in classification phase on meta set of training data, then results of ensemble of predictors is fused into single consensus prediction. The consensus module, namely MLcons meta-learner, is the core part of brainstorming approach. Each ML is used independently to predict class membership for a query object. The consensus is then build using results from all representations and features describing this object by different learners in prediction phase. The final output includes predicted class membership, statistical model with performances of each learning modules, trained consensus and reliability scores for prediction.

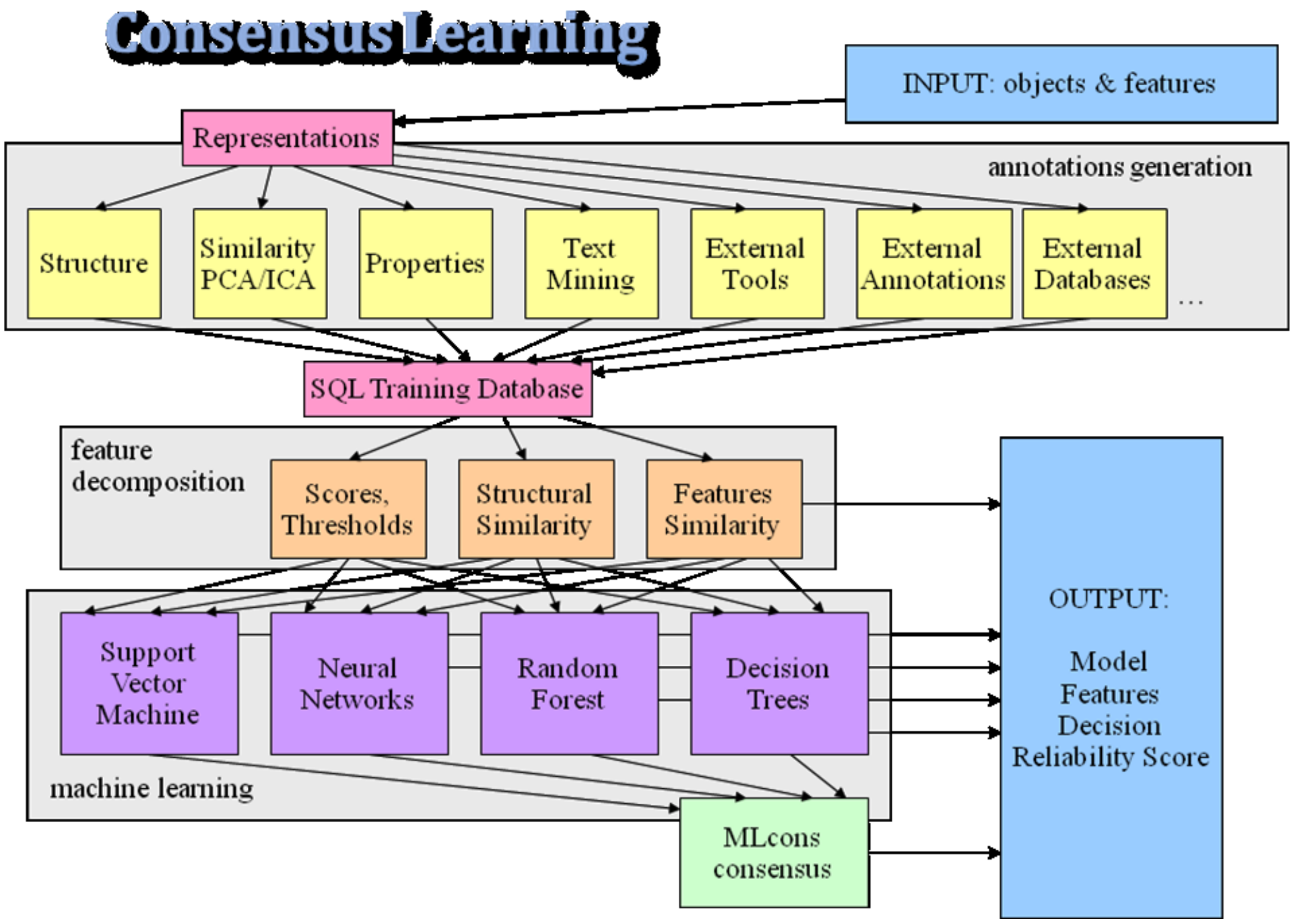
Consensus Learning
INPUT: objects & features
Representations
annotations generation
Structure
Similarity PCA/ICA
Properties
Text Mining
External Tools
External Annotations
External Databases
...
SQL Training Database
feature decomposition
Scores, Thresholds
Structural Similarity
Features Similarity
Support Vector Machine
Neural Networks
Random Forest
Decision Trees
machine learning
MLcons consensus
OUTPUT:
Model
Features
Decision
Reliability Score